(Preprint)

# GUIDANCE, NAVIGATION AND CONTROL OF MULTIROBOT SYSTEMS IN COOPERATIVE CLIFF CLIMBING

Himangshu Kalita,[*] Ravi Teja Nallapu,[†] Andrew Warren,[‡] and Jekan Thangavelautham [§]

The application of GNC devices on small robots is a game-changer that enables these robots to be mobile on low-gravity planetary surfaces and small bodies. Use of reaction wheels enables these robots to roll, hop, summersault and rest on precarious/sloped surfaces that would otherwise not be possible with conventional wheeled robots. We are extending this technology to enable robots to climb off-world canyons, cliffs and caves. A single robot may slip and fall, however, a multirobot system can work cooperatively by being interlinked using spring-tethers and work much like a team of mountaineers to systematically climb a slope. A multirobot system as we will show in this paper can climb surfaces not possible with a single robot alone. We consider a team of four robots that are interlinked with tethers in an "x" configuration. Each robot secures itself to a slope using spiny gripping actuators, and one by one each robot moves upwards by crawling, rolling or hopping up the slope. If any one of the robots loses grip, slips or falls, the remaining robots will be holding it up as they are anchored. This distributed controls approach to cliff climbing enables the system to reconfigure itself where possible and avoid getting stuck at one hard to reach location. Instead, the risk is distributed and through close cooperation, the robots can identify multiple trajectories to climb a cliff or rugged surface. The benefits can also be realized on milligravity surfaces such as asteroids. Too fast a jump can result in the robot flying off the surface into space. Having multiple robots anchored to the surface keeps the entire system secure. Our work combines dynamics and control simulation to evaluate the feasibility of our approach. The simulation results show a promising pathway towards advanced development of this technology on a team of real robots.

**INTRODUCTION**

Wheeled and legged robots have been studied extensively in the recent years for exploration of extreme environments. Some of the legged robots can even climb and maneuver on vertical surfaces. However, they are still limited from exploring extreme environments such as caves, lava tubes and skylights in off-world environments like the Moon or Mars due to inherent challenges in motion planning and control on dusty surfaces. We follow a different approach to solving this problem by utilizing teams of fully autonomous robots that hop, perform short flights and roll[1]. These missions may require traversing low-gravity surfaces of asteroids, bypassing impassable terrains or climbing extremely rugged terrains such as canyons, cliffs, craters walls and caves to acquire critical science data (Figure 1). Exploring these off-world terrains is daunting and it requires a holistic systems solution that utilizes the latest in robotic mobility combined with smart planning to recover from missteps and slips. Guidance Navigation and Control devices such as reaction-wheels, IMUs together with a propulsion system enables unprecedented mobility in precarious surface conditions. We have proposed SphereX, a spherical robot, 3 kg in mass, and 30 cm in diameter that can hop, fly, roll and summersault on planetary surfaces and small-bodies. Moreover, with the addition of a suitable gripping skin, these robots can grasp onto rough terrain and rest on precarious/sloped surfaces. Hence, these robots can climb up a slope by hopping/rolling a distance $d$ and then gripping on the surface. However, a single robot may slip and fall if the gripping mechanism fails to grasp. This can be avoided by developing a multirobot system that can work cooperatively by being interlinked using spring-tethers and work much like a team of alpine mountaineers to systematically climb a slope.

---

[*] PhD Student, Space and Terrestrial Robotic Exploration Laboratory, Arizona State University, 781 E. Terrace Mall, Tempe, AZ.
[†] PhD Student, Space and Terrestrial Robotic Exploration Laboratory, Arizona State University, 781 E. Terrace Mall, Tempe, AZ.
[‡‡] Undergraduate Student, Space and Terrestrial Robotic Exploration Laboratory, Arizona State University, 781 E. Terrace Mall, Tempe, AZ.
[§] Assistant Professor, Space and Terrestrial Robotic Exploration Laboratory, Arizona State University, 781 E. Terrace Mall, Tempe, AZ.

In this paper, we present dynamics and control simulation of an autonomous multirobot system that cooperates to climb sloped surfaces by successively hopping, rolling and crawling. A multirobot team exceeds the sum of its parts by tackling complex slopes that would otherwise be too risky for a single robot to traverse. The multirobot system comprises of four spherical robots that are interlinked with tethers in an "x" configuration. Each robot is secured to a slope using spiny gripping actuators, and one by one each robot moves upwards by crawling, rolling or hopping up the slope. If any one of the robots loses grip, slips or falls, the remaining robots will be holding it up as they are anchored.

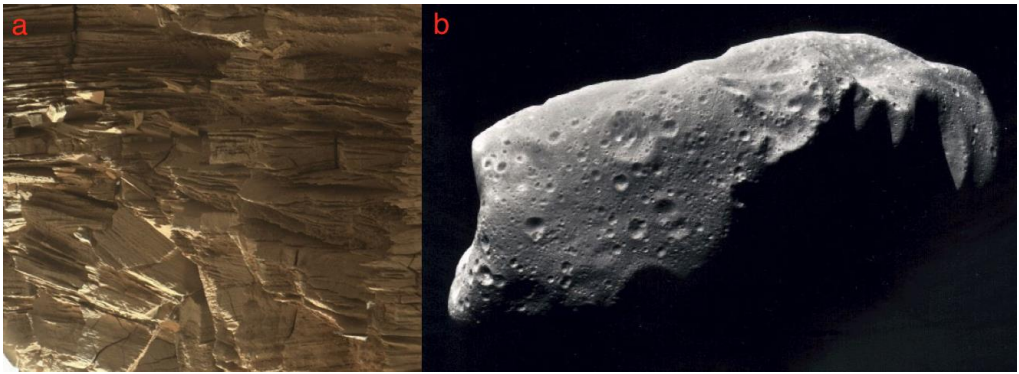

**Figure 1. (Left) Cliff faces on Mars. (Right) Asteroid 2009 ES.**

This multirobot approach for climbing sloped cliff surfaces holds great potential for exploring cliff and extremely rugged surface environment on Mars, Moon and asteroids. Recent research suggests that water flowed down the faces of several Martian cliffs as seen in high-resolution images acquired by the Mars Global Surveyor Orbiter Camera[2]. Getting up-close, traversing down these slopes enables going back in time to better under the geological history of Mars. These extreme environments cannot be accessed using conventional wheeled, legged or rolling robots. Hence, there is an important need to develop next-generation robotic systems that can reach these sites by flying or climbing steep slopes.

Moreover, the benefits of the SphereX system can be realized on milligravity surfaces such as asteroids. There are 150,000+ asteroids, with a large number located in the asteroid belt between Mars and Jupiter[23]. They range in size with diameters ranging from a few meters to several hundred kilometers. On milligravity surfaces, hopping and flying is simple and uses negligible propellant. However, the gravity varies throughout the surface and too much thrust can result in a spacecraft achieving escape velocity. Using the proposed multirobot approach with robots anchored to the surface keeps the entire system secure. In the following sections, we present background and related work followed by system overview, dynamic simulations, discussions, conclusions and future work.

**RELATED WORK**

Climbing remains a major challenge in robotics. Much work has focused on developing tethered legged and wheeled robots. Dante II is an eight-legged walking rover that was used to explore the craters of volcanoes using a tethered rappelling mobility system[3]. However, it was not a fully autonomous system and required teleoperation. Another example is the All-Terrain Hex-Limbed Extra-Terrestrial Explorer (ATHLETE) rover developed by NASA JPL[4]. ATHLETE has six 6-DOF limbs, each attached with a 1-DOF wheel. The wheels can be used for efficiently driving over smooth terrains and it can be locked and used as feet to overcome steep obstacles or rugged terrains. Another example is the Teamed Robots for Exploration and Science on Steep Areas (TRESSA) that was used for climbing steep cliff faces with slopes varying from 50 to 90 degrees[5]. It is a dual-tethered system that allows lateral motion on steep slopes and successfully demonstrated semi-autonomous science investigations of cliffs. Another example is Axel developed by NASA JPL which is a two-wheeled rover tethered to its host platform for enhancing mobility on challenging terrains like steep slopes and overhangs[6]. It is capable of in-situ measurements and sampling on challenging terrains and successfully demonstrated accessing 90 degree vertical cliffs and collecting samples.

The Legged Excursion Mechanical Utility Rover (LEMUR IIb) developed by NASA JPL is a four-limbed robot that can free-climb vertical rock surfaces[7]. In addition to vertical rock surfaces, it can traverse a variety of other terrains like urban rubble piles, sandy terrain and roads using only friction at contact points. Several climbing robots employing suction cups, magnets and sticky adhesives. One such example is the Stickybot developed at Stanford that employs several design principles adapted from the gecko lizard like hierarchical compliance, directional adhesion and force control to climb smooth surfaces at very low speeds[8]. Another robot developed is Spinybot II that can climb a wide variety of hard, outdoor surfaces including concrete, stucco, brick and sandstone by employing arrays of microspines that catch on surface irregularities[9]. The Robots in Scansorial Environments (RiSE) is a new class of vertical climbing robots that can climb a variety of human-made and natural surfaces employing a combination of biologically inspired attachments, dynamic adhesion and microspines[10]. Another application of micro spines developed by NASA JPL has been an anchoring foot mechanism for sampling on the surface of near Earth asteroids[11]. The mechanism can withstand forces greater than 100 N on natural rock and has been proposed for use on the Asteroid Retrieval Mission (ARM).

Another technique developed at Stanford and NASA JPL uses 3-axis reaction wheels to creep over rugged surfaces no matter how steep or uneven[27]. The technique works well in low-gravity and small but steep slopes. It is unclear how a gyroscopic system will handle large steep surfaces; as they are bound to slip and fall after extend use/missteps. Multirobot systems have been tested for space applications including exploration; base-preparation and resource-mining[18]. The motivation for our multirobot system is taken from proven methods used by alpinists to climb mountains. These mountaineers use ice axes and crampons to grip on the surface and climb steep mountain slops as shown in Figure 2. The use of legs and hands provide four contact points to the sloped surface. Even when each attempt to grip onto a higher location fails, the climber is still secure with his feet and one hand gripping tightly onto the slope.

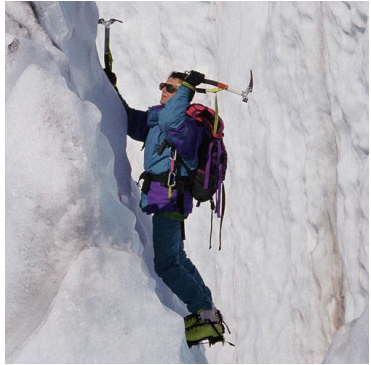

**Figure 2. Mountain climbers use an ice axe and crampons to climb steep icy slopes. They use their two hands and two feet to grip onto the icy slope.**

Inspired by mountaineers, our approach utilizes a multirobot climbing and flying system that has inherent redundancies to recover from individual missteps and slips. Our proposed approach is a total systems solution to address the challenge of off-world climbing. The system utilizes multiple SphereX robots that are interlinked with spring-tethers. The multirobot system works cooperatively to fly over impassable terrain and climb off-world cliffs, canyons and caves. The system uses an array of microspines to grip on the rough surface while climbing.

**SYSTEM OVERVIEW**

Our proposed design consists of four spherical robots interlinked together with spring tethers in an "x" configuration. Figure 3 shows the internal and external views of each spherical robot, without the micro-spine skin. The lower half of the sphere contains the power and propulsion system, with storage tanks for fuel and oxidizer connected to the main thruster. The attitude control system is at the center and contains a 3-axis reaction wheel system for maintaining roll, pitch and yaw. The main thruster enables translation along the $+z$ axis and in combination with the attitude control system it enables the robot to move along 3-axes. Next is the Lithium Thionyl Chloride batteries with specific energy of 500 Wh/kg arranged in a circle as shown.

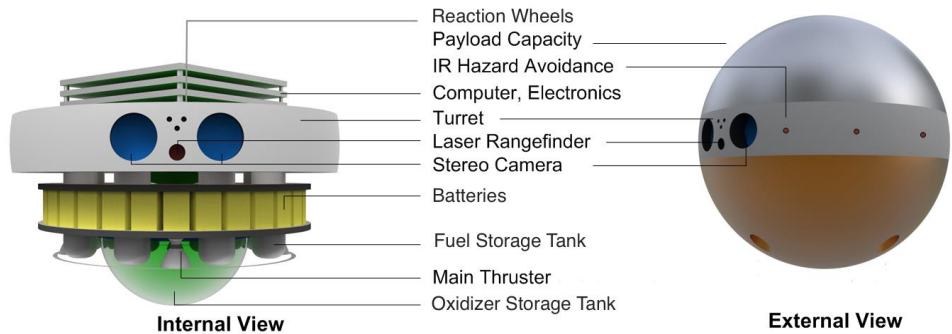

**Figure 3. Internal and external views of each SphereX robot.**

An alternative to batteries are PEM fuel cells. PEM fuel cells are especially compelling as techniques have been developed to achieve high specific energy, solid-state fuel storage systems that promise 2,000 Wh/kg[15,16,17]. However, PEM fuel cells require further development for a field system in contrast to lithium thionyl chloride that has already been demonstrated on Mars.

A pair of stereo cameras and a laser range finder rolls on a turret. This enables the robot to take panoramic pictures and scan the environment without having to move using the propulsion system. Moreover, the stereo camera and laser range finder would aid in navigation and perception. Above the turret are two computer boards, IMU and IO-expansion boards, in addition to a power board. The volume above the electronics is reserved for climbing mechanism and equipment of up to 1 kg[1].

Apart from the proposed propulsion subsystem, all other hardware components can be readily assembled using Commercial Off-the-Shelf (COTS) CubeSat components. The proposed propulsion system uses RP-1 as the fuel and $H_2O_2$ as the oxidizer. The mass budget for a single SphereX robot is shown in Table 1.

**Table 1. SphereX robot mass budget**

| Major Subsystem | Mass (kg) |
| --- | --- |
| Computer, Comms, Electronics | 0.2 |
| Power | 0.3 |
| Stereo Camera, Laser Rangefinder | 0.3 |
| Propulsion | 0.8 |
| ADCS | 0.4 |
| Climbing Payload | 1 |
| Total | 3 |

**DYNAMIC SIMULATION**

We are using four SphereX robots, each of mass 3 kg. Each robot has a propulsion unit to provide the thrust required for hopping and a 3-axis reaction wheel system to change the orientation of the robot. The combination of propulsion unit and the reaction wheels will help us achieve ballistic hopping capabilities. The surface area of each robot consists of a microspine that enables the robot to grip onto rough surfaces while climbing sloped cliffs.

**Ballistic Hops using Rocket Propulsion**

For ballistic hops, a liquid propellant rocket motor is used to provide thrust along the +z axis. Analysis has been done for different types of solid and liquid propellants based on their $I_{sp}$, flight time and feasibility[14]. The thrust generated by a rocket motor depends on the mass flow rate, nozzle exhaust velocity and combustion chamber pressure as shown below[12]:

$$F = A_{th} p \sqrt{\frac{2k^2}{k-1}\left(\frac{2}{k+1}\right)^{\frac{k+1}{k-1}}\left[1 - \left(\frac{p_e}{p}\right)^{\frac{k-1}{k}}\right]} \quad (1)$$

where, $F$ is the thrust generated, $p$ is the combustion chamber pressure, $A_{th}$ is the nozzle throat area, $p_e$ is the nozzle exit pressure and $k$ is the ratio of specific heats. With the thrust provided by the rocket motor along +z axis, a set of reaction wheel is used to control the orientation of the ro-bot which enables the robot to move along 3-axes. The reaction wheel system applies torque to the spherical robot about its principal axes according to the control command resulting in change in orientation and angular velocity. This is done by applying conservation of angular momentum to the robot and reaction wheel system, and can be expressed by setting the time derivative of the total angular momentum to zero shown below[13]:

$$\vec{\dot{L}} = J_B \cdot \vec{\dot{\omega}_B} + \vec{\omega_B} \times (J_B \cdot \vec{\omega_B}) + J_{RW} \cdot \vec{\dot{\omega}_{RW}} + \vec{\omega_B} \times (J_{RW} \cdot \vec{\omega_{RW}}) = 0 \quad (2)$$

where, $L$ is the angular momentum of the system, $J_B$ and $J_{RW}$ are the moment of inertia of the robot and the reaction wheels, $\omega_B$ and $\omega_{RW}$ are the angular velocity of the robot and reaction wheels respectively. The system has three inputs that are torques exerted by the reaction wheels. The three outputs are the desired Euler angles of the robot. The required torques can be calculated by a PD control algorithm as shown below:

$$\tau_{rw} = -K_p(e_{des}\text{-}e_{act}) - K_d(\omega_{des}\text{-}\omega_{act}) \quad (3)$$

where, $\tau_{rw}$ is the torque generated by the reaction wheels, $K_p$ and $K_d$ are the proportional and derivative controller gains, $e_{des}$ and $e_{act}$ are the desired and actual Euler angles, $\omega_{des}$ and $\omega_{act}$ are the desired and actual angular velocity of the spherical robot respectively. Figure 4 shows the trajectory of a spherical robot for a PD control algorithm in a Martian environment with acceleration due to gravity of 3.71 m/s². The desired Euler angles were 0.27, 0.25 and 0.07 radians and the desired angular velocities were 0 rad/s. Moreover, we have simulated the system with RP-1 as a fuel and hydrogen peroxide ($H_2O_2$) oxidizer and it consumes 5 grams of propellant for every hop. It is clear from the figure that for every hop, the robot can travel a distance of 0.37 m along x-axis, 0.41 m along y-axis and can attain a height of 0.28 m along z-axis (Figure 4).

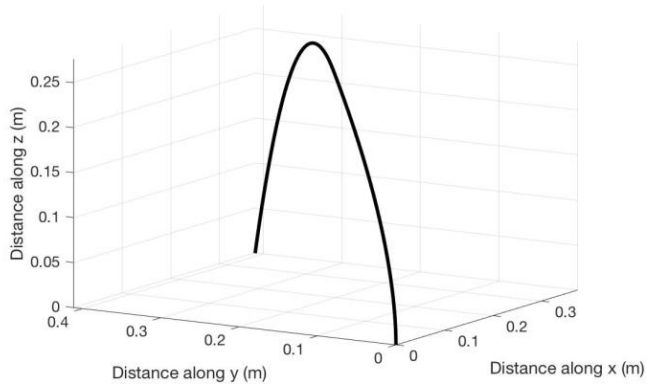

**Figure 4. Trajectory of SphereX robot performing a rocket-propelled ballistic hop.**

**Ballistic Hops using only Reaction-Wheels**

The ballistic hopping mechanism discussed above, uses a liquid propellant rocket motor and a system of 3-axis reaction wheel. Although this mechanism can achieve controlled hopping, the system does expend significant amounts of fuel. An alternative approach to hopping utilizes orthogonal reaction wheels surrounded by external spikes[19,20,21] as shown in Figure 5.

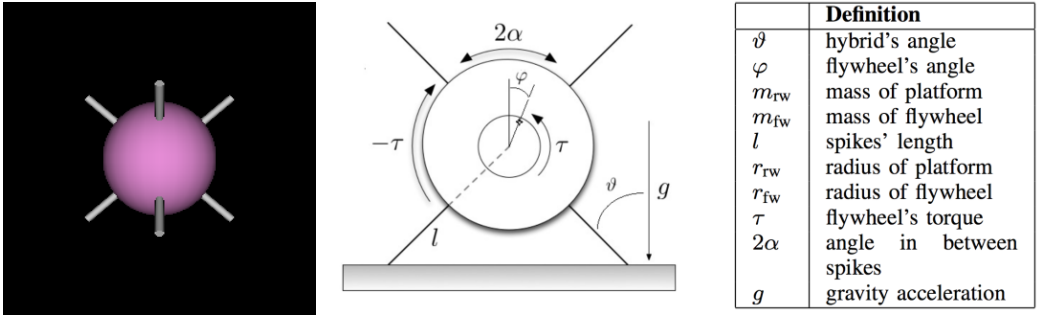

**Figure 5. Externa layout of robot using reaction wheels to perform ballistic hops.**

Using a linear combination of internal torques with the help of the three reaction wheels, the system can produce momentarily large reaction forces at the surface. The contact forces between the spikes and the surface comprises of spring-damper forces normal to the surface, and a Coulomb friction component tangential to the surface[21,22]. With sufficient torque applied, the system can produce momentary large reaction forces, causing the platform to leave the surface, hopping forward in a ballistic trajectory. The approach consists of a hybrid control algorithm, where the reaction wheels are slowly accelerated to a desired angular velocity, and the impulsively braked to generate the torque needed to produce hopping[20]. With this control strategy, the desired angular velocity $\omega$ and braking torque $\tau$ can be regarded as the two control variables and they are a function of lateral distance to be covered $d$ and acceleration due to gravity $g$.

Figure 6 shows the desired angular velocity and braking torque required to hop a lateral distance of 1 m as a function of $g$. For our analysis, we have considered the mass of each reaction wheel as 0.35 kg, radius of each reaction as 3.5 cm, length of each spike from the center of the robot as 25 cm. For surfaces like Phobos with acceleration due to gravity 0.006 m/s$^2$, the desired angular velocity of the reaction wheel is 314 rad/s (~3,000 rpm) and the desired braking torque is 0.063 Nm. However, for Mars, it is 7,952 rad/s (~75,940 rpm) and 38.8 Nm respectively which is extremely high and not practical.

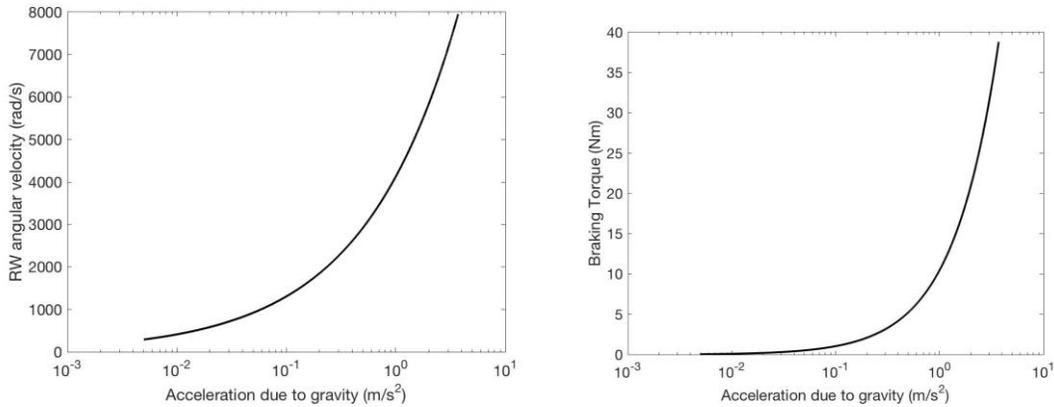

**Figure 6. Angular Velocity (left) and braking torque (right) to hop a lateral distance of 1 m as a function of $g$.**

Figure 7 shows the lateral hopping distance, $d$ as a function of input torque, $t$ and reaction wheel speed $w$ on a surface with acceleration due to gravity 0.006 m/s$^2$. It shows that a SphereX robot can hop up to a distance of 4 m with an input torque of 0.33 Nm and reaction wheel speed of 6,000 rpm. This makes reaction-wheel based ballistic hops practical for milligravity environments.

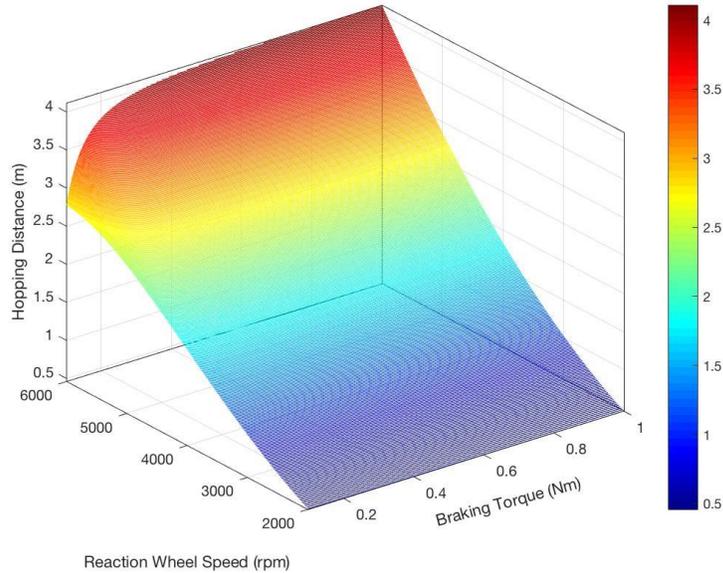

**Figure 7. Later hopping distance as a function of reaction wheel input torque and reaction wheel speed surface with acceleration due to gravity 0.006 m/s$^2$ (Phobos).**

**Gripping Mechanism**

The gripping mechanism consists of microspines. Each microspine toe consists of a steel hook embedded in front of a rigid frame with elastic flexures acting as a suspension system (Figure 8)[11]. For a spine of tip radius $r_s$, it will engage to asperities of average radius $r_a$ such that $r_a \geq r_s$.

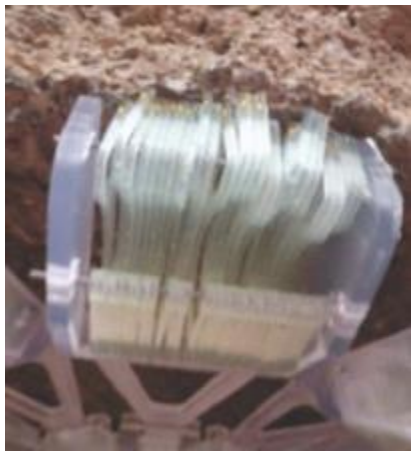

**Figure 8: Microspine toe securely gripping and hanging from a rocky surface[11].**

Engagement of the spine to asperities depend on the angle $\theta$ of the normal vector to the traced surface and is possible only if it is larger than some critical angle $\theta_{min}$. The angle $\theta_{min}$ depends on the angle at which the spines are loaded, $\theta_{load}$, and coefficient of friction, $\mu$, between the steel hook and the rocky surface as shown below[9]:

$$\theta_{min} = \theta_{load} + \cot^{-1} \mu \tag{4}$$

Hence, smaller spines with smaller tip radius $r_s$ are more effective at engaging to asperities on smooth surfaces. However, smaller spines also carry smaller loads. Moreover, the maximum load of the spine/asperity contact increases as $r_s^2$, while the expected number of asperities per unit area decreases as $1/r_s^2$. Thus, as we decrease the tip radius of the hook, it can engage to smoother asperities but the load carrying capacity decreases[9]. The elastic flexures act as a suspension system and allow each hook to move relative to its neighbors. When the array of microspines are dragged along a surface each toe is stretched and dragged to find a suitable asperity to grasp and share the overall load uniformly. This system of microspines can attach to both convex and concave asperities as shown in climbing robots like RiSE[10] and Spinybot[9]. The maximum load that a spine can sustain is a function of the tensile stress of the hook and square of the radii of curvature of the spine tip and asperity as shown is below[9].

$$f_{max} = \left[\left(\pi \sigma_{max}/(1-2\mu)\right)^3 (1/2E^2)\right] R^2 \tag{5}$$

$$\text{where, } \frac{1}{R} = \frac{1}{r_s} + \frac{1}{r_a} \tag{6}$$

where, $\sigma_{max}$ is the tensile stress and $E$ is the modulus of elasticity of the material.

**Cliff Climbing Multirobot System**

For climbing sloped surfaces, each spherical robot is equipped with an array of microspines. The robot hops using the propulsion system and reaction wheels and then grips on the rough surface using the array of microspines. However, climbing sloped or vertical cliffs for a single robot is a risky matter. A single robot may slip and fall if the gripping mechanism fails to grasp into the rough surface. However, a multirobot system can work cooperatively by being interlinked using spring-tethers and work much like a mountaineer to systematically climb a slope. We have considered a system of four spherical robots that are interlinked with four spring-tethers in an "x" configuration which work cooperatively to climb a slopped rough surface as shown in Figure 9.

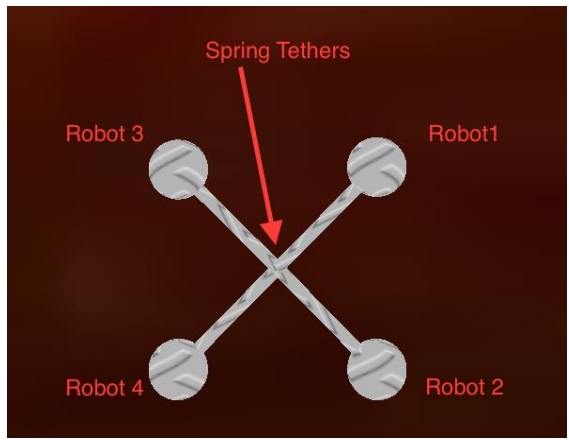

**Figure 9. Cliff climbing multirobot system.**

The connections between the robots and tether are made with ball and socket joints. The spring tether introduces one translational degree of freedom in the system which allows each robot to translate with respect to the other robots. The ball-socket joint introduces three rotational degrees of freedom in the robot-tether connection and the tether-tether connection, which allows each robot to hop with respect to the other robots resulting in 3-dimensional movement of the whole system. Figure 10 shows a Matlab 3D VRML dynamics simulation of a team of 4 robots climbing a slopped surface. In Figure 10.1 all robots are gripping onto the slope surface. Robot 1 disengages its grip and hops a distance *d* forward and then grips again on the surface. When robot 1 hops, the other three robots are still gripped to the surface, hence if robot 1 loses grip, slips or falls, the remaining robots will be holding it up as they are anchored. Robot 1 continues to hop until it is able grip onto the surface at a distance *d* from its initial position. Similarly, in Figure 10.3-10.5 robot 2, 3 and 4 hops and grips on the surface as shown until each robot is displaced by a distance *d*. Figure 10.5 shows the final configuration of the robot system after it had climbed a distance *d* up the slope.

The surface of each SphereX robot consists of hundreds of microspines. For the robot to climb a wide variety of rough surfaces, it has a combination of large spines as well as smaller spines spread uniformly. Each spine has a shaft diameter of 200-300 µm and a tip radius of 12-25 µm. The maximum load that each spine can sustain per asperity is 1-2 N. Each robot has a mass of 3 kg and each tether has a mass of 0.15 kg, making the mass of the whole system approximately 12.6 kg. On Mars, with a *g* of 3.7 m/s$^2$, the spines need to sustain a load of 47 N. With each spine/asperity contact capable of sustaining 1-2 N load, a minimum of 28 spines should be engaged. With each robot rolling or hopping at a time, the other three robots must share the total load, hence a minimum of 10 spines need be engaged for each robot.

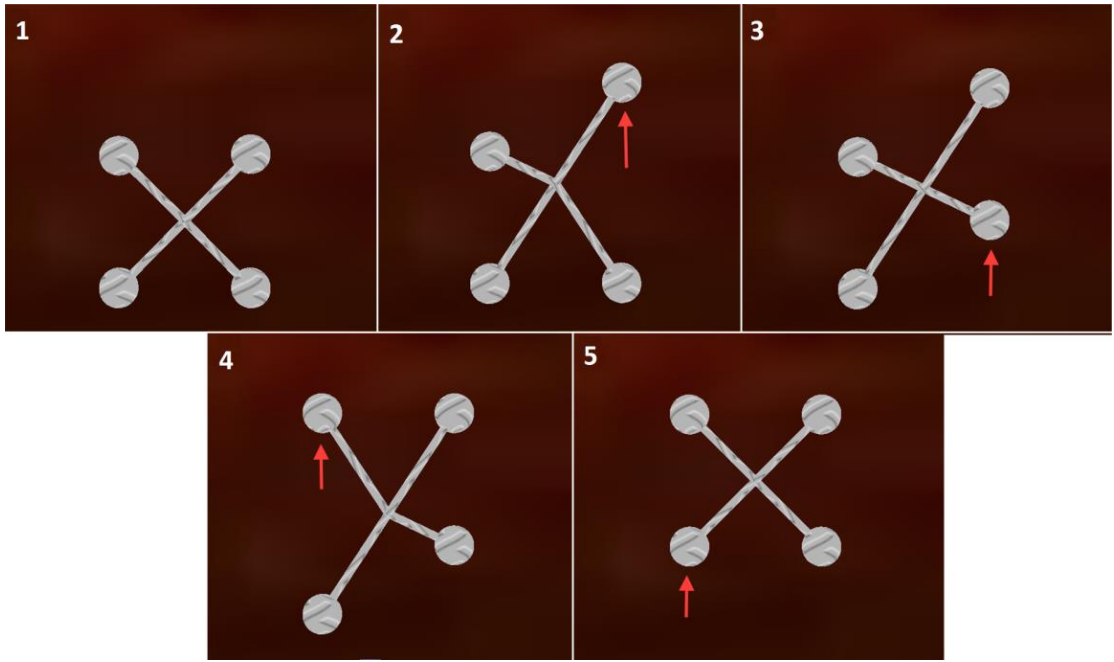

**Figure 10. Sequence of robot movement to climb a steep slope. Each robots hops up the slope, individually and in sequence and grips to the surface. The robots are all attached using spring tether.**

Figure 11 shows how the position of each robot and the "instantaneous center" of the whole system changes with time. The initial position of robot 4 is at the origin (0,0,0) and that of robot 1, robot 2 and robot 3 are (1,1,0), (1,0,0) and (0,1,0) (m) respectively. It is clear that each robot hops one at a time resulting in the change in position of the instantaneous center. Figure 12 shows the change in $x$, $y$ and $z$ coordinates of the "instantaneous center" of the system as its climbing. After four successive hops total, one by each robot, the instantaneous center moves a distance of 0.75 m along $y$-axis in 10 seconds.

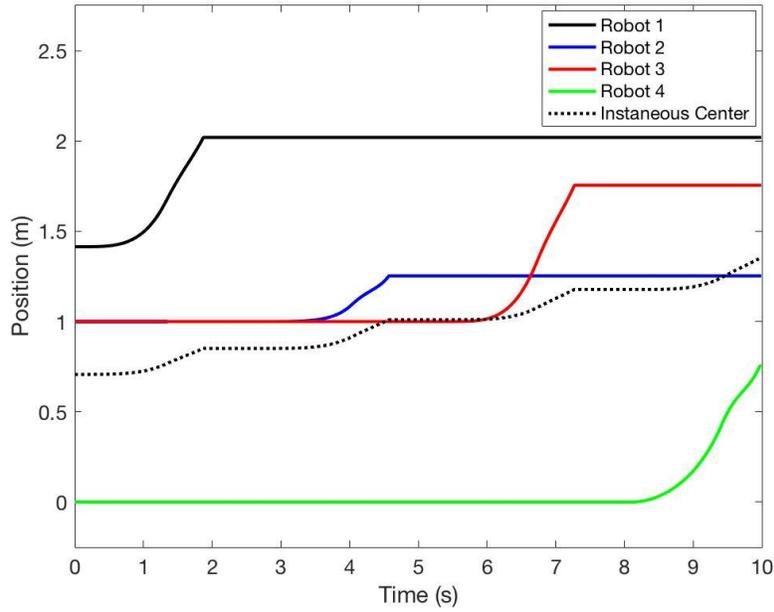

**Figure 11. Change in position of each robot and the instantaneous center during a climb.**

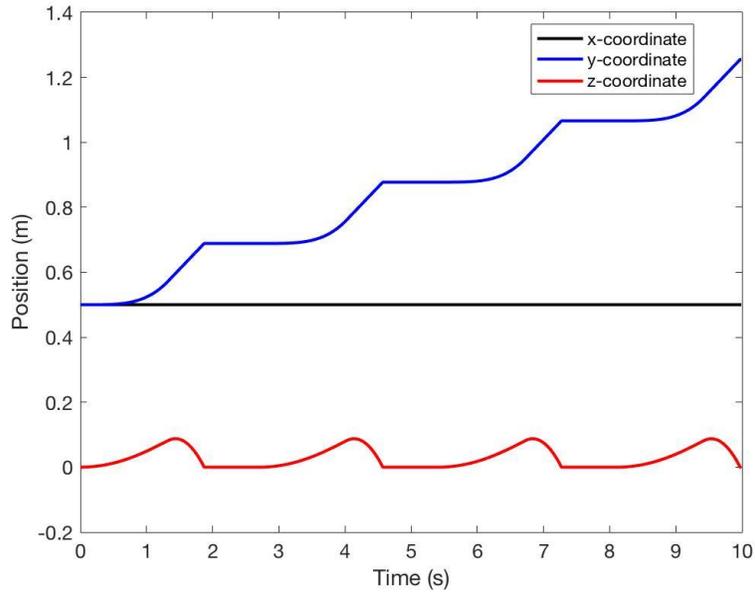

**Figure 12. Change in x, y and z coordinate of the instantaneous center during a climb.**

Figure 13 shows the y-coordinates of each robot and the "instantaneous center" when robot 1 fails to grip on the surface after hopping. All the robots successfully grip onto the surface after the first hop. Then, robot 1 fails to grip after its second hop and slips down. However, the remaining robots hold it up as they are anchored. Robot 1 then hops again to attain the desired height and grips on the surface on its next attempt. Failure of any robot to grip on a surface leads to more consumption of fuel and time to climb a certain height.

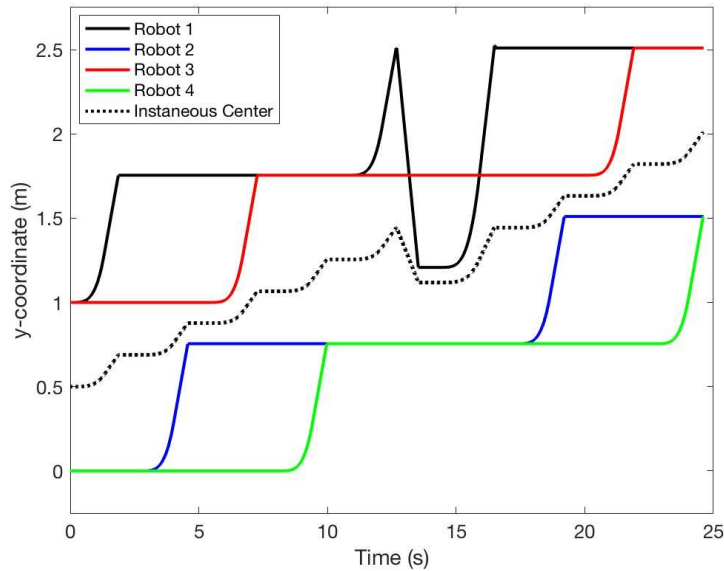

**Figure 13. Y-coordinate of each robot and the instantaneous center when robot 1 fails to grip on the surface.**

**DISCUSSION**

The proposed cooperative cliff climbing technique using four spherical robots interlinked with tethers in an "x" configuration is suitable for exploring cliff faces on Mars, the Moon, surfaces of asteroids and other planetary bodies. With an array of microspines attached to each robot, the multirobot system can grip to any rough surface and then climb or crawl without the risk of falling from a cliff or flying off an asteroid. Moreover, each robot has hopping/flying capability with the help of the propulsion and the ADCS system. Climbing enables persistent access of the sloped surface. This multirobot system has unique advantage over other wheeled or legged climbing robot systems.

This multirobot system can have higher climbing speed compared to other wheeled or legged systems due to the use of a propulsion system. Each robot can hop a distance of 0.75 m in 2 seconds on Mars with an expenditure of 5 grams of RP1-$H_2O_2$ propellant. Alternative methods to hop include use of a mechanical hopping mechanism and use of reaction-wheels applied with a braking torque. With four robots interlinked the system must perform four successive hops to climb a particular distance. Assuming each robot can grip onto the surface on its first attempt, the whole system can climb a distance of 0.75 m in approximately 10 seconds. In milligravity surfaces like asteroids, the climbing/crawling speed will be even more efficient. The robots can use the reaction wheels alone to hop on an asteroid. In addition, the robots could simply fly to a location and land onto to the side of the cliff before performing climbing a few meters to reach a desired science target. The system of stereo cameras and laser rangefinder enables the robots to accurately navigate the surface and find the best possible path to climb avoiding obstacles.

**CONCLUSION**

In this paper we have introduced a multirobot system for cooperative cliff climbing and steeped surface exploration. This multirobot system is suitable for climbing cliff faces on the Moon, Mars and other low-gravity bodies. We have devised a system that can withstand individual missteps, slips or falls by a robot during the climbing process. A combination of flying, hopping and climbing enables the proposed system to access hard to reach sites. However, the system does expend significant quantities of fuel when using propulsion. Alternative approaches to hopping include use of reaction wheels applied with braking-torque. This approach enables hopping without use of propellant however it is only feasible for low-gravity environments such as surface of asteroids and small-bodies. The dynamics and control simulations for a single hopping robot with propulsion and ADCS system were presented. Finally, the cliff climbing mechanism was simulated using four robots and four spring tethers. The paper presents insight on the feasibility and the advantages/disadvantages of this multirobot system for exploring steeped planetary surfaces, asteroids and small-bodies.